\documentclass{article}

\usepackage{microtype}
\usepackage{graphicx}
\usepackage{subcaption}
\usepackage{booktabs} 

\usepackage{hyperref}

\usepackage[preprint]{icml2026}

\usepackage{amsmath}
\usepackage{amssymb}
\usepackage{mathtools}
\usepackage{amsthm}
\usepackage{multirow,multicol}

\usepackage[capitalize,noabbrev]{cleveref}

\theoremstyle{plain}

\theoremstyle{definition}

\theoremstyle{remark}

\DeclareFixedFont{\ttb}{T1}{txtt}{bx}{n}{12} 
\DeclareFixedFont{\ttm}{T1}{txtt}{m}{n}{12}  

\usepackage{color}
\definecolor{deepblue}{rgb}{0,0,0.5}
\definecolor{deepred}{rgb}{0.6,0,0}
\definecolor{deepgreen}{rgb}{0,0.5,0}

\usepackage{listings}
\usepackage[frozencache]{minted}

\newcommand\pythonstyle{\lstset{
language=Python,
basicstyle=\ttm,
morekeywords={self},              
keywordstyle=\ttb\color{deepblue},
commentstyle=\color[HTML]{228B22}\sfffamily,
emph={MyClass,__init__},          
emphstyle=\ttb\color{deepred},    
stringstyle=\color{deepgreen},
breaklines=true,
columns=fullflexible,
frame=tb,                         
showstringspaces=false,
}}

\lstnewenvironment{python}[1][]
{
\pythonstyle
\lstset{#1}
}
{}

\newcommand\pythoninline[1]{{\pythonstyle\lstinline!#1!}}

\usepackage[textsize=tiny]{todonotes}

\icmltitlerunning{Finding a Generalizable Activation Functions}

\usepackage{listings}

\begin{document}

\twocolumn[
  \icmltitle{Mining Generalizable Activation Functions}



  \icmlsetsymbol{equal}{*}

  \begin{icmlauthorlist}
    \icmlauthor{Alex Vitvitskyi}{gdm}
    \icmlauthor{Michael Boratko}{gdm}
    \icmlauthor{Matej Grcic}{gdm}
    \icmlauthor{Razvan Pascanu}{gdm}
    \icmlauthor{Deep Shah}{gdm}
    \icmlauthor{Petar Veli\v{c}kovi\'{c}}{gdm}
  \end{icmlauthorlist}

\icmlaffiliation{gdm}{Google DeepMind}
 \icmlcorrespondingauthor{Alex Vitvitskyi}{avlife@google.com}

  \icmlkeywords{Activation function, evolutionary search, out of distribution generalization}

  \vskip 0.3in
]



\printAffiliationsAndNotice{\icmlEqualContribution}

\begin{abstract}
 The choice of activation function is an active area of research, with different proposals aimed at improving optimization, while maintaining expressivity. Additionally,  the activation function can significantly alter the implicit inductive bias of the architecture, controlling its non-linear behavior. In this paper, in line with previous work, we argue that evolutionary search provides a useful framework for finding new activation functions, while we also make two novel observations. The first is that modern pipelines, such as AlphaEvolve, which relies on frontier LLMs as a mutator operator, allows for a much wider and flexible search space; e.g., over all possible python functions within a certain FLOP budget, eliminating the need for manually constructed search spaces. In addition, these pipelines will be biased towards meaningful activation functions, given their ability to represent common knowledge, leading to a potentially more efficient search of the space. The second observation is that, through this framework, one can target not only performance improvements but also activation functions that encode particular inductive biases. This can be done by using performance on out-of-distribution data as a fitness function, reflecting the degree to which the architecture respects the inherent structure in the data in a manner independent of distribution shifts. We carry an empirical exploration of this proposal and show that relatively small scale synthetic datasets can be sufficient for AlphaEvolve to discover meaningful activations.
 \end{abstract}

\section{Introduction}

The hidden layer of a neural network is the basic building block for modern AI systems. It consists of a learnable linear projection followed by a non-linear \emph{activation function}, typically applied element-wise, which enables the stacking of hidden layers to lead to an increase in expressivity \cite{montufar14neurips,raghu17icml}. This structure is meant to mimic a rudimentary understanding of the biological neurons, where information flows through synaptic connections, whose strength is meant to be represented by the parameters contained within the linear projections. The activation function not only reflects the non-linear behavior of 
biological neurons, but allows for composing (``stacking'') of several hidden layers without the result collapsing to a single linear projection. 

However, the parallel between artificial neural networks and biological systems is limited \cite{crick89nature}. 
For example, biological systems rely on spike trains \cite{roy19nature}, whereas neural networks rely on continuous-valued signals, allowing the use of gradient based optimization techniques. Therefore, biology can only go so far in terms of indicating what is the correct parametrization of a hidden layer, and in particular, what is the right choice of non-linearity. 

This realization led to an extensive search for non-linearities in the AI community~\citep[e.g][]{Ramachandran2018SearchingFA}, where various activation functions have been explored \cite{Dubey2022-hp}, from hand-crafted activations such as $\operatorname{ReLU}$ \cite{nair10icml}, hyperbolic tangent \cite{leCun12}, $\operatorname{GELU}$ \cite{hendrycks16arxiv}, and so forth, all the way to explicit automated search procedures. This exploration was driven by the ability of the final architecture to obtain 
better \emph{in-domain} performance on standard benchmarks. The improved performance, 
as shown in various works, comes from a matching between initialization and the choice 
of activation function, where most benefits come from better behaved gradients and better
signal propagation in the model, rather than expressivity. 

In this work, we perform an \emph{evolutionary} search for activation functions, in a way that aims to \emph{transfer} performance from carefully-constructed small-scale experiments to improved downstream out-of-domain generalisation.  

\section{Contributions}

\begin{figure*}[ht]
  \vskip 0.2in
  \begin{center}
    \centerline{\includegraphics[width=.8\textwidth]{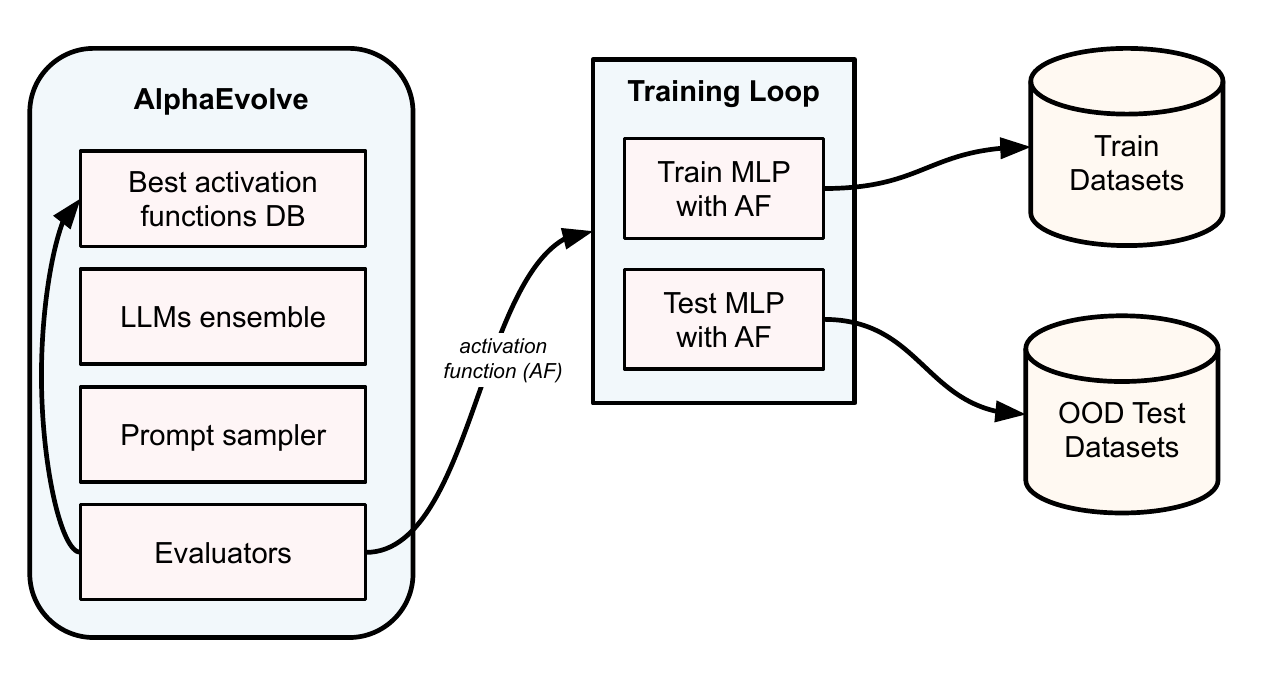}}
    \caption{
      Overall description of the evolutionary search framework we use to find activation functions. We rely on an evolutionary search powered by AlphaEvolve \citep{novikov25preprint}, designed to optimize test performance for small-scale models trained on carefully constructed, synthetic datasets. We demonstrate that functions discovered in this way are capable of meaningful forms of generalization without sacrificing their general-purpose potency.
    }
    \label{icml-historical}
  \end{center}
\end{figure*}

We depart from previous approaches in two important ways. First, rather than evolving using function composition over a pre-defined set of simpler operations ~\citep[e.g.][]{Ramachandran2018SearchingFA}, we explore an \emph{unbounded} set of functions. To navigate in this infinite space of functions, we rely on the AlphaEvolve framework \cite{novikov25preprint}, which operates over the space of all Python programs. A key component in this approach is the reliance on frontier language models (e.g. Gemini \citep{comanici2025gemini}) to generate meaningful proposals during search. If desired, these models can also be guided using expert information in the prompt, as a way of injecting additional biases into the search. We will describe our specific setup in the following sections. 

In contrast to previous works~\citep{Ramachandran2018SearchingFA}, we explicitly target \emph{validation} loss on \emph{out-of-domain data} as our fitness function. The rationale for this decision is to attempt to generate activation functions that allow for better \emph{generalization}, either through their implicit impact on the loss landscape or training dynamics.
While functions discovered in this way will typically not capture explicit symmetries or structures in the data exactly, their ability to target and improve on out-of-domain performance demonstrates the flexibility of our approach.

Moreover, our empirical results demonstrate that while standard known 
activation functions are generally powerful and hard to beat on in-distribution performance, our approach can discover alternatives which exhibit better out-of-domain generalization without sacrificing the in-domain performance. By targeting performance on out-of-domain validation data, our search results in architectural choices with improved robustness to unseen data.

While it is theoretically possible to directly search for activation functions or even the entire architecture layout that optimise larger training runs on massive datasets -- typically explored within the domain of neural architecture search (NAS) \citep[e.g.][]{zoph2017neural, elsken2019neuralarchitecturesearchsurvey} -- this can be computationally expensive, and may lead to overfitting to a particular dataset.

Instead, we propose a simpler protocol, which we can refer to as the \emph{small-scale lab} approach, which amounts to optimizing activation functions within \emph{small} networks on small-scale synthetic data, allowing for rapid iteration of the inner loop of the evolutionary search strategy. This ensures that we can easily train such models from scratch every time we need to compute the fitness value of every proposal. By focusing on out-of-domain generalization, and ensuring the chosen synthetic evaluation data is sufficiently complex, we aim for the discovered activations to straightforwardly apply in more complex settings and help with improving generalization performance. In work concurrent to ours, \citet{nadimpalli2025evolving} use classical reinforcement learning environments as another highly relevant testbed; however, they do not explicitly study out-of-domain generalization, and such settings scarcely occur in classical RL tasks.

Once we have mined several activation functions using our approach, we evaluate them within established baseline architectures on standard datasets, such as CIFAR-10 \citep{krizhevsky2009learning}, ImageNet \citep{deng2009imagenet}, CLRS-30 \citep{velivckovic2022clrs} and ogbg-molhiv \citep{hu2020open}. Datasets such as CIFAR-10 and ImageNet typically do not test for an explicit distribution shift---as opposed to CLRS-30, which directly tests for size generalization. The ogbg-molhiv task also attempts to test for a distribution shift, in that it uses an explicit \emph{scaffold split} of its molecular data, requiring generalization to new structures. We find that our approach is able to discover simple activation functions which significantly improve CLRS-30 generalization compared to established baselines, while not penalizing performance on tasks such as ImageNet.

\section{AlphaEvolve for activation functions}

AlphaEvolve \cite{novikov25preprint} is an evolutionary coding system, powered by a set of large language models (LLMs).  The main components of AlphaEvolve are samplers (which propose new solutions), evaluators (which provide feedback on solution quality), and a database of the best solutions proposed so far. In our specific case, AlphaEvolve will aim use its evaluator to incrementally improve the generalization performance of the activation functions produced by the sampler.

\paragraph{Search}
Specifically, our instantiation of AlphaEvolve will perform its search by following these steps:
\begin{enumerate}
    \item Look-up a set of previously proposed activation functions (initially, we seed this process with only the standard activation function $\mathrm{ReLU}(x) = \max(0, x)$).
    \item Conditioned on the code of the recovered functions, use frontier LLMs to generate new function implementation proposals, aimed towards improving the performance of the model.
    \item The generated proposals are evaluated by instantiating small-scale multilayer perceptron (MLP) architectures with the proposed activation functions, and training them on a series of purpose-built synthetic datasets. 
    \item The trained models are then evaluated on out-of-distribution test data, and the computed evaluation errors are used as the fitness function for the evolutionary process (deciding which functions to keep in the set of proposals for future iterations of the system).
    \item Iterate steps 1--4 until satisfactory functions are found.
\end{enumerate}
\begin{figure}
    \centering
    \includegraphics[width=\linewidth]{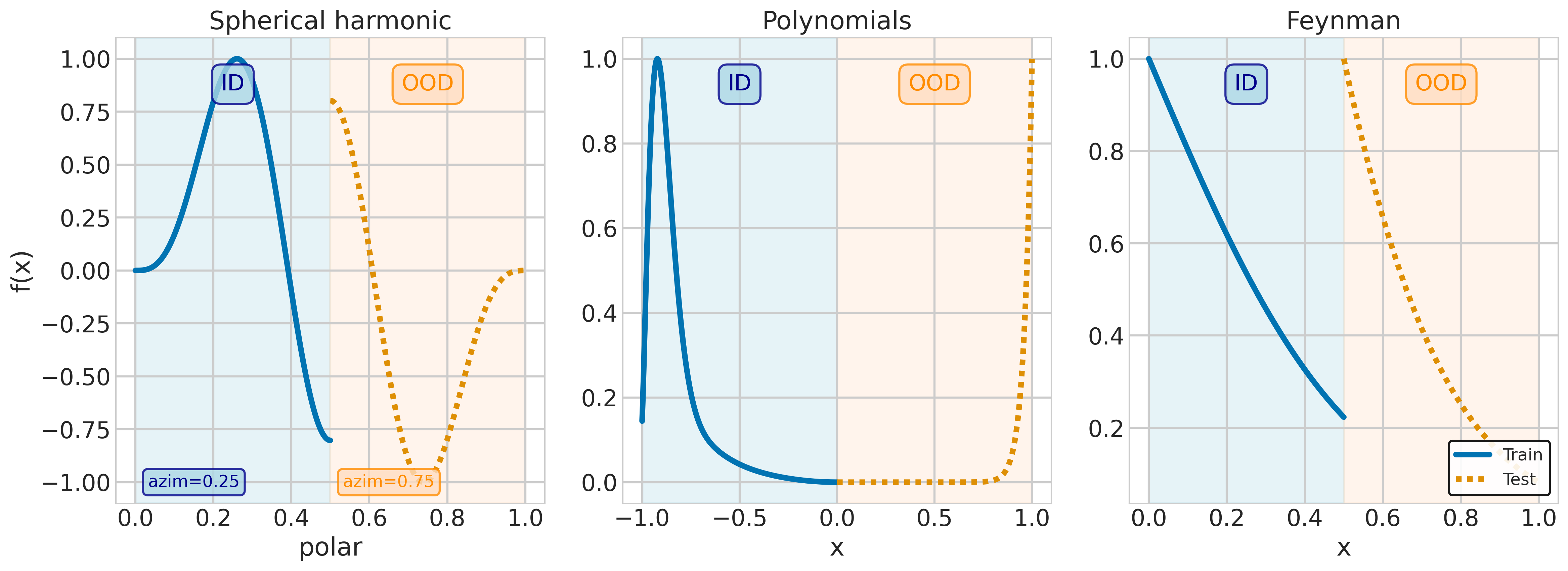}
    \caption{Visualization of in-distribution (training) and out-of-distribution (test) data for the one-dimensional target functions leveraged in our small-scale lab environment. Each function type tests a different kind of generalization in a way that supports rapid model training. Note that ID and OOD panels are \emph{different} functions---there are no discontinuities in the functions we study.}
    \label{fig:train_fn}
\end{figure} 
\begin{figure*}
    \includegraphics[width=\linewidth]{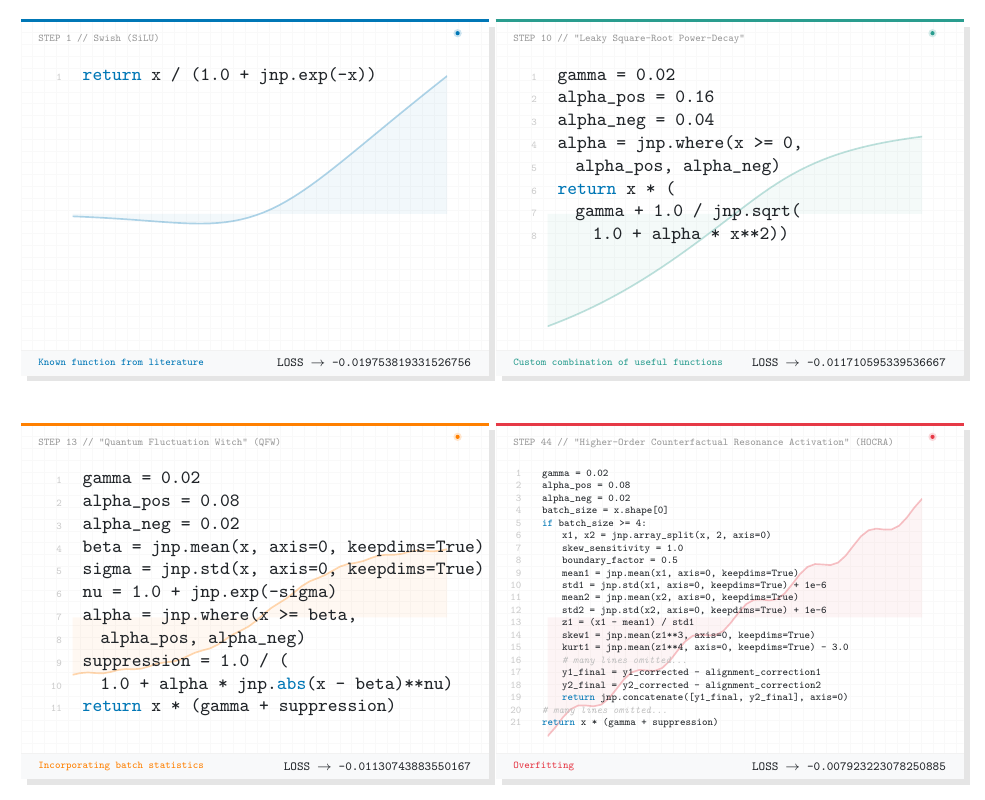}
    \caption{Illustration of a typical AlphaEvolve evolution when asked to discover new activation functions. Early on, it rediscovers functions present in the literature (Swish/SiLU in this case). It then discovers interesting ways to recombine standard building blocks (polynomials, leaky ReLU, and square roots), at which point it reaches the best tradeoff between score and transferability. Soon after, AlphaEvolve realises that its function does not need to be pointwise, and leverages the batch axis of the input tensor to extract and exploit basic batch statistics. This quickly spirals into constructing highly elaborate functions that achieve excellent score, but heavily overfit to the specifics of the ``lab dataset'' by utilising multiple moments of the distribution.}
    \label{fig:activation_evolution}
\end{figure*}
\paragraph{Datasets} For our purpose-built training/evaluation data, we focus on synthetic regression datasets that are designed to test various aspects of out-of-domain generalization. Concretely, our synthetic datasets are input-output examples of the form $(x, f(x))$, where the target function, $f$, is applied to random low-dimensional input data, $x \in \mathbb R^d$. 

Our collection of target functions $f$ includes the following:

\begin{itemize}
    \item Polynomials with random coefficients;
    \item Harmonic functions;
    \item The Feynman Symbolic Regression Dataset \citep{udrescu2020ai}.
\end{itemize}

Figure \ref{fig:train_fn}
visualizes the examples of such functions that were used for MLP training and evaluation within the AlphaEvolve loop.
Each plot clearly indicates in-distribution training data (in blue) and out-of-distribution test data (in orange). These targets are designed to support different kinds of generalizing behaviour -- from finding parameters that extrapolate well (polynomials), to being able to encode data with structural regularity (harmonic/periodic functions), all the way to being generally capable on a variety of physics-inspired equations (Feynman lectures).

All of these are simple functions that can be learned relatively quickly in-distribution on modern hardware, allowing for fast evaluation. To ensure our test data is sufficiently out-of-domain from the training data, we generate random input points over disjoint intervals: train on the range $(0,0.5)$, test on the range $(0.5,1)$ or on the range $(0,1)$, test on the range $(-1,0)$. 

\paragraph{Evaluation}
We use the negative mean squared error (MSE) as the scoring function for AlphaEvolve, which allows for iterative improvements of the activation function. In addition, we promote efficiency in the generated functions by placing a hard limit on the number of permitted floating-point operations (FLOPs), so that the computational cost of training with the proposed activation functions would not change significantly from the baseline choice of $\mathrm{ReLU}$. 

\begin{figure*}[htbp]
\centering
\includegraphics[width=\textwidth, keepaspectratio]{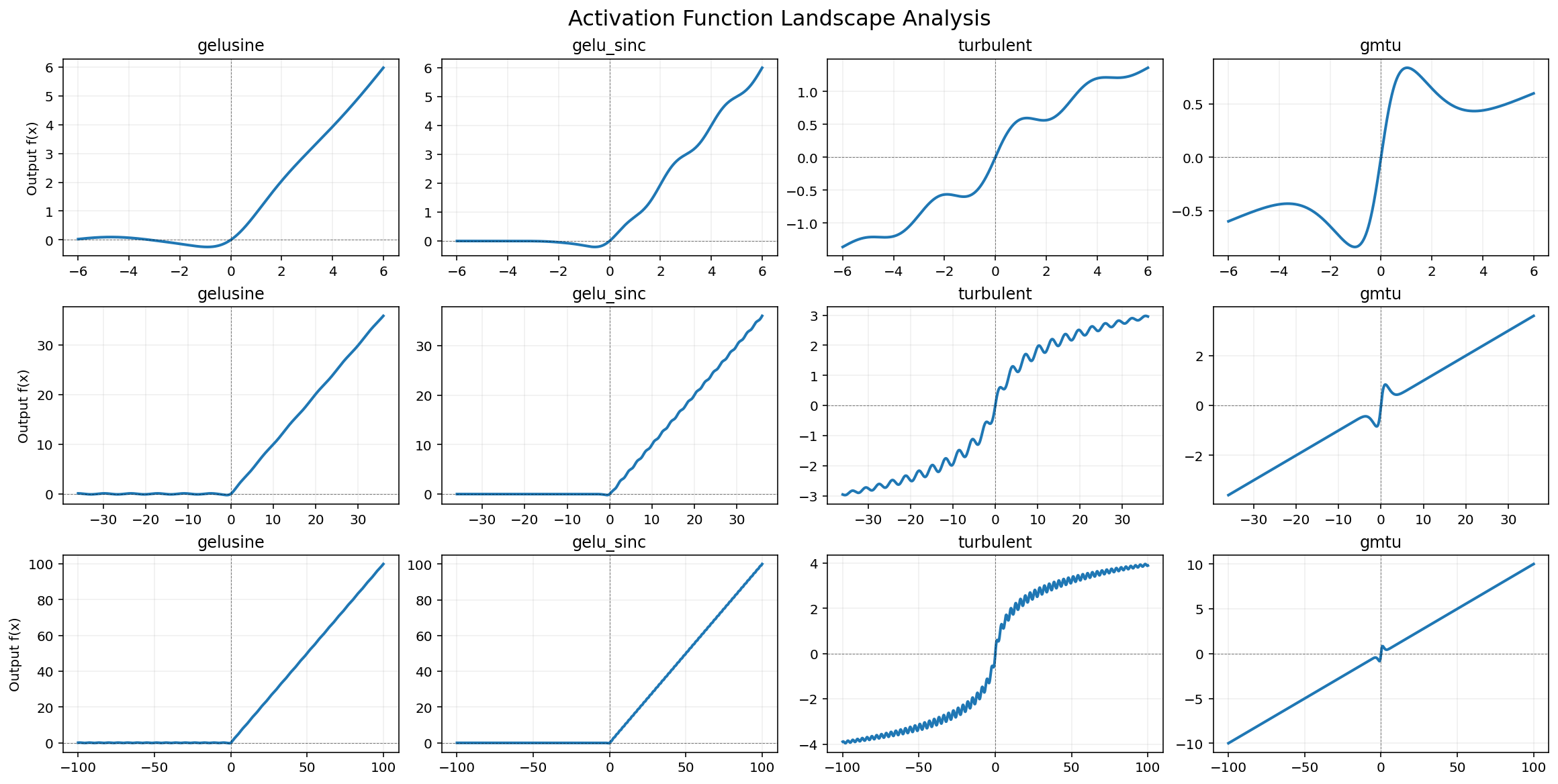}
\caption{Visualisation of four newly discovered pointwise activation functions by our system: \emph{Turbulent Activation Function}, \emph{Gaussian-Modulated Tangent Unit (GMTU)}, \emph{GELUSine} and \emph{GELU-Sinc-Perturbation}.}
\label{fig:activ_functions}
\end{figure*}

AlphaEvolve discovers suitable activation functions iteratively, usually starting from re-discovering simple well-known activation functions (such as $\mathrm{GELU}$ \citep{hendrycks16arxiv}), gradually allowing for more complex combinations of such functions, and ultimately coming up with very elaborate functions which explicitly inspect the batch axis of the data. Such functions leverage the statistical properties of the complete tensor of activations across a batch, making them more prone to overfitting to statistics present in the synthetic data. As such, it is necessary to identify the ``sweet spot'' in the evolutionary approach, or otherwise restrict the evolutionary engine for relying too much on such correlations. See Figure~\ref{fig:activation_evolution} for a depiction of these key points within a function evolution run.

\begin{table}
    \centering
    \caption{Overall performance of discovered activation functions by AlphaEvolve on the synthetic generated data, comprising Polynomials, Harmonics, and Feynman symbolic regression.}
    \label{tab:my_results}
    \small
        \begin{tabular}{lcc} 
        \toprule
        \multirow{2}{*}{\textbf{Activation Function}} & \textbf{Test Loss} & \textbf{Train Loss} \\
                                                      & \textbf{($\times 10^{-3}$)} & \textbf{($\times 10^{-3}$)}  \\
        \midrule
        Turbulent & \textbf{29.8}             & 3.5 \\
        GMTU      & 51.9    & 4.1 \\
        GELUSine        & 54.7            & \textbf{1.4} \\
        GELU-Sinc  &68.7	& 2.2  \\
        GELU &78.8	&2.0 \\ \midrule
        ReLU &93.1	& 3.5 \\
        \bottomrule
    \end{tabular}
\end{table}

\begin{figure*}[htbp]
\centering
\includegraphics[width=\textwidth, keepaspectratio]{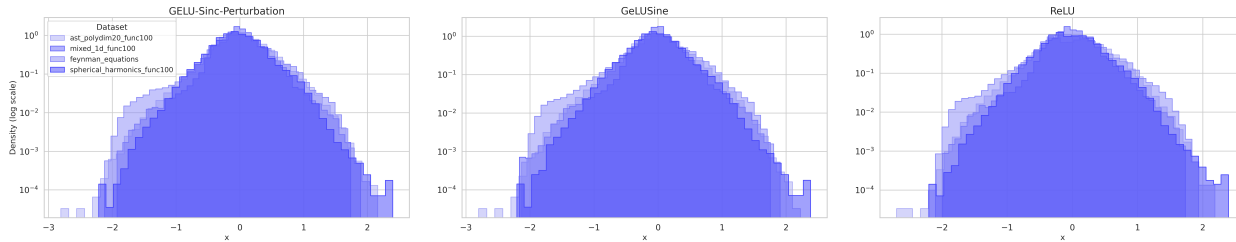}
\caption{Histogram of the pre-activation entries for trained model on synthetic dataset for:\emph{GELUSine} and \emph{GELU-Sinc-Perturbation} and \emph{ReLU}. Note the range tends to be wide enough, i.e. $[-2,2]$, allowing the model to exploit the structure of the activation functions. }
\label{fig:preactiv_hist}
\end{figure*}

While we use synthetic datasets for the initial discovery of promising activation functions, we then validate the generalization abilities of the top-performing candidates by replacing default activations in established baseline architectures with the proposed functions. These models are then trained from scratch on standard datasets, including CIFAR-10, CLRS-30 and ogbg-molhiv. This allows us to validate our discoveries, and find activation functions that provide good generalization capabilities independent of our choice of synthetic training datasets.

\section{Activation function analysis}

We dedicate the remainder of our work to analysing further some of the most promising activation functions discovered by our small-scale lab. We initially select nine such functions that had the most promising performance on our collection of synthetic tasks (and include our initial function, $\mathrm{ReLU}$, as a baseline). 

\begin{table*}
\small
\caption{Activation Functions Comparison}\label{tab:complete_results}
\begin{tabular}{llrrlrrrr}
\toprule
 & CIFAR  & CLRS  & molhiv  & ImageNet   & Feynman  & 1d poly  & 20d poly  & Sph. H. \\
 & $\uparrow$  top\_1 & $\uparrow$ test & $\uparrow$ test & $\uparrow$  top\_1 & $\downarrow$ test & $\downarrow$ test & $\downarrow$ test & $\downarrow$ test \\
Activation Function & accuracy & score & AUC & accuracy & MSE & MSE & MSE & MSE \\
\midrule
Gaussian-Modulated Tangent Unit (GMTU) & 0.915 & 0.861 & 0.784 & 0.676 & 0.053 & 0.104 & 0.015 & 0.035 \\
Gaussian Error Linear Unit (GELU) & 0.948 & 0.874 & 0.758 & 0.745 & 0.056 & 0.195 & 0.056 & 0.008 \\
GELUSine & 0.946 & 0.867 & 0.765 & 0.745 & 0.043 & 0.138 & 0.033 & 0.005 \\
GELU-Sinc-Perturbation & 0.948 & 0.887 & 0.776 & 0.739 & 0.043 & 0.160 & 0.052 & 0.020 \\
Turbulent Activation Function & 0.886 & 0.833 & 0.755 & 0.610 & 0.024 & 0.071 & 0.018 & 0.006 \\
Quaternion-Inspired & 0.514 & 0.888 & 0.699 & OOM & 0.006 & 0.006 & 0.004 & 0.004 \\
Fourier-Informed Spectral Gating (FISG) & 0.345 & 0.894 & 0.655 & 0.167 & 0.002 & 0.002 & 0.008 & 0.001 \\
Phase-Locked Entropic Repulsion & OOM & 0.891 & OOM & OOM & 0.009 & 0.001 & 0.019 & 0.000 \\
Symmetric Phase-Flipped & 0.229 & 0.878 & 0.657 & 0.001 & 0.002 & 0.001 & 0.004 & 0.001 \\ \midrule
ReLU & 0.947 & 0.862 & 0.756 & 0.735 & 0.030 & 0.101 & 0.162 & 0.019 \\
\bottomrule
\end{tabular}
\end{table*}

Out of these functions, we mainly focus on four that are newly discovered by our system -- \emph{turbulent}, \emph{Gaussian-Modulated Tangent Unit (GMTU)}, \emph{GELUSine} and \emph{GELU-Sinc-Perturbation} -- their code may be found in Appendix \ref{app:codes}. All of the functions are pointwise except for the turbulent activation, which computes simple batch statistics. As previously described, we expect pointwise functions to be more generally reusable compared to functions computing batchwise statistics; further, they are much easier to visually analyse (see Figure \ref{fig:activ_functions}). 

Table~\ref{tab:my_results} contains the overall performance scores of these four functions on the synthetic datasets described in the previous section, in the form of the training and test loss they achieve. We also include two standard baselines -- the starting point of our search, the rectified linear unit ($\operatorname{ReLU}$) and the Gaussian error linear unit ($\operatorname{GELU}$), which is frequently rediscovered by our search.  

In terms of the training loss on the synthetic tasks, $\operatorname{GELU}$ is competitive with the discovered functions, aligning well with previous observations in the activation search literature. Specifically, it is well understood that standard performant activation functions often converge in similar ways on training data. The critical differentiator is generalization on out of distribution test data: the previously-known functions ($\operatorname{GELU}$ and $\operatorname{ReLU}$) are worse than all discovered activation functions, with the \emph{turbulent} function yielding a markedly better test loss than all competitors, despite its training loss being identical to that of $\operatorname{ReLU}$.

\begin{figure*}
    \centering
    \includegraphics[width=\linewidth, keepaspectratio]{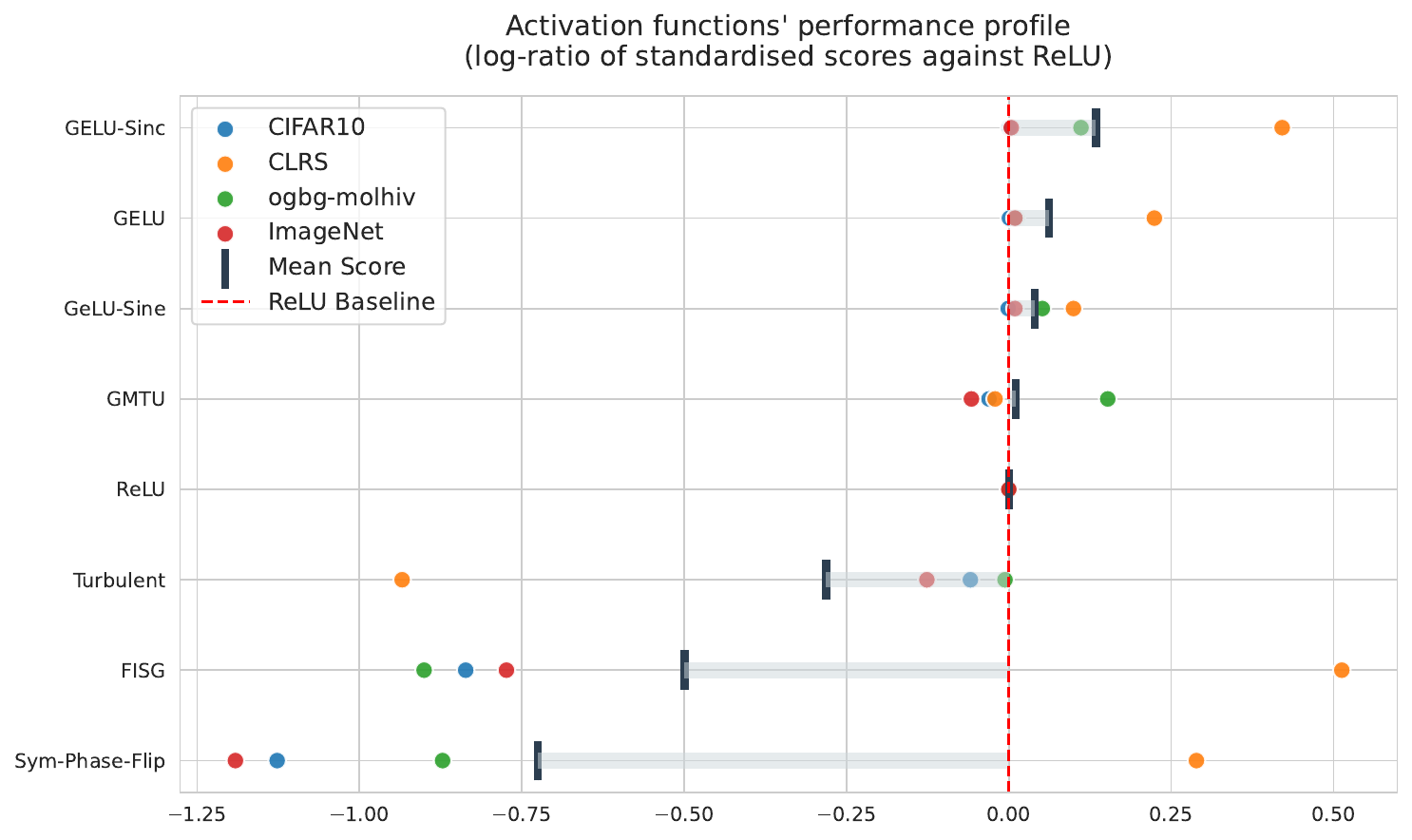}
    \caption{Performances of the discovered activation functions, plotted against the ReLU baseline, on four downstream tasks (CIFAR-10, CLRS-30, ogbg-molhiv and ImageNet).}
    \label{fig:overall_results}
\end{figure*}

The other interesting observation that can be made inspecting these functions---both from their implementations in Appendix \ref{app:codes} and plots in Figure \ref{fig:activ_functions}---is that they all have shapes related to established activation functions. Specifically, they usually take the form of the sum or product of two terms. The first term tends to have the form of a standard activation function, e.g. $\operatorname{GELU}$, while the second term is usually a function that has a periodic shape and modulates this standard shape in a way that might be better at fitting periodic patterns. Two key examples are the GELUSine:
\begin{equation}
    \mathrm{GELUSine}(x) = \mathrm{GELU}(x) + 0.1\sin x
\end{equation}
and the GELU-Sinc-Perturbation:
\begin{align*}
    \mathrm{GELUSinc}(x) &= \mathrm{GELU}(x)(1 + 0.5\ \mathrm{sinc}\ x)\\ &= \mathrm{GELU}(x)\left(1 + \frac{\sin(\pi x)}{2\pi x}\right)
\end{align*}
In both of these cases, we note the usage of sine waves, which might allow a model to store information on in-domain data which would then get more easily retrieved when going out-of-domain, its reappearance guaranteed by the sine function's periodicity.

In Figure~\ref{fig:preactiv_hist}, for \emph{GELU-Sinc-Perturbation} and \emph{GeLUSine} as well as $\operatorname{ReLU}$,
we visualise the distribution of the pre-activation entries, which on the synthetic data ranges from $[-2,2]$. This confirms that the activation functions are not used in a limited range where they would exhibit a very similar shape as typical function, but rather the pre-activation are large enough to take advantage of the additional structure of the activation function.

\begin{figure}
    \centering
    \includegraphics[width=\linewidth]{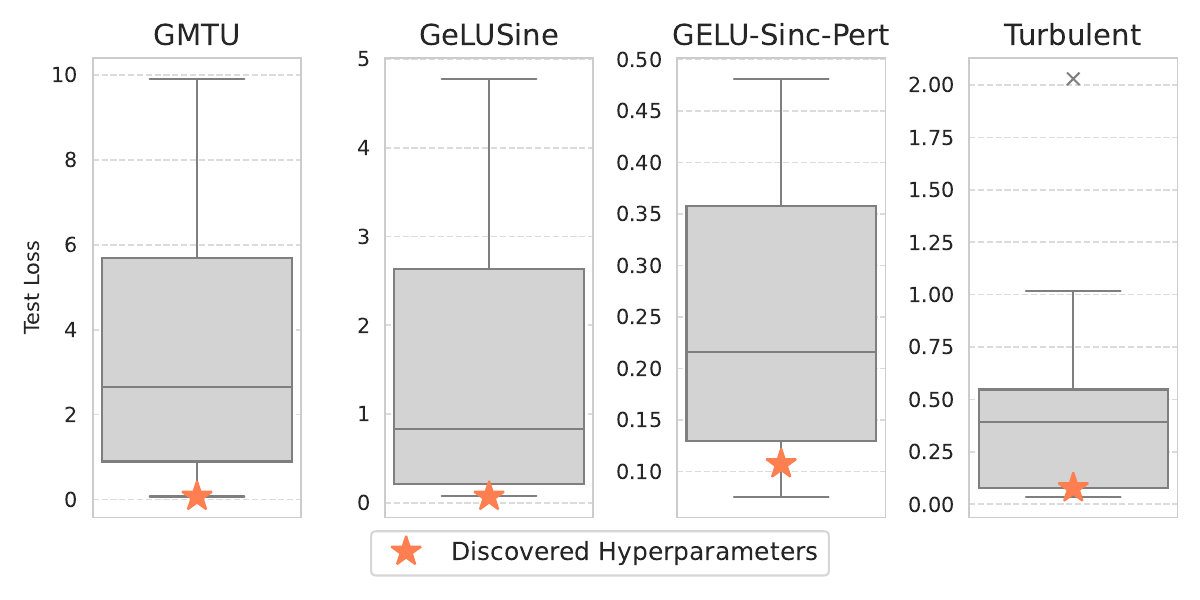}
    \caption{AlphaEvolve discovers hyperparameter configuration that outperform randomly-selected function combinations of the same form on our small-lab test datasets.}
    \label{fig:hparam_sensitivity}
\end{figure}

As it appears that function combinations are important for generalization, how can we reason about the optimal combinations between two given functions, and has AlphaEvolve found them? We attempt to answer this by recognizing that all of the multiplicative constants used in these expressions can be replaced by a configurable hyperparameter, $\alpha$, and we can then randomly sweep many sensible values of $\alpha$ -- see Appendix \ref{app:ablation_gelusine} for the specific case of GELUSine. The result of this experiment may be observed in Figure \ref{fig:hparam_sensitivity}, illustrating that the specific values of $\alpha$ proposed by AlphaEvolve (e.g. $0.1$ for GELUSine or $0.5$ for GELUSinc) generally outperform the majority of the random samples drawn from our hyperparameter grid. 

\section{Downstream evaluation}

We now reintroduce the complete set of ten functions under study---including functions that tend to explicitly take into account batch statistics, such as the \emph{Quaternion-inspired}, \emph{FISG}, \emph{Phase-Locked Entropic Repulsion} and \emph{Symmetric Phase-Flipped} activation. Their implementations may also be found in Appendix \ref{app:codes}. The complete test results across both our small-scale lab and the downstream tasks may be observed in Table \ref{tab:complete_results}, and summarized results relative to our ReLU baseline on the downstream tasks---only for functions that did not cause any out-of-memory issues---may be observed in Figure \ref{fig:overall_results}.

The tasks selected for this analysis (CIFAR-10, CLRS-30, ogbg-molhiv and ImageNet) do not necessarily share any commonalities with the synthetic datasets we use; the goal of our evaluation is to ensure that the evolution process did not overfit to the small-scale synthetic data, and to assess whether any meaningful out-of-domain generalization has emerged on downstream tasks that require it (in this case, we specifically target CLRS-30, as it has an explicit OOD test dataset). 

It may be readily observed that all functions exploiting the batch axis severely underperform on the image-based datasets, even if they may occasionally provide OOD benefits on the CLRS dataset. This implies that they have sacrificed their general applicability to achieve stronger generalisation on ``raw'' distribution shifts. In some cases, the batch statistics computations even result in out-of-memory (OOM) issues. Additionally, it is notable that the \emph{turbulent} activation does not generalise well to any new dataset; in spite of its superior pointwise performance in the lab, it has unfortunately learnt an overly-fitted combination of standard functions, and failed to transfer outside of the lab.

On the other side of the performance spectrum, we note that the (re-discovered) \emph{Gaussian Error Linear Unit (GELU)}, as well as its (newly-discovered) variants (\emph{GELUSine} and \emph{GELU-Sinc-Perturbation}) all outperform the ReLU baseline. This implies a solid degree of transfer outside of the lab, and validates our approach of using small scale synthetic datasets to discover activation functions that appear to be generically applicable in different task instances. 

In particular, the novel \textbf{GELU-Sinc-Perturbation} function appears to perform the best overall -- offering a differentiating performance on the CLRS-30 benchmark, a representative task requiring OOD generalization, without sacrificing performance on the image benchmarks, and offering improvements on the molecular benchmark ogbg-molhiv. 

This discovery implies that it is possible to usefully transfer findings outside of the small-scale lab we built and into larger architectures. Furthermore, it indicates that simple compositions of known performant activations with carefully-calibrated periodic components may unlock useful gains in generalization ability, while retaining the generality and utility of the base function. In future work, we believe it would be important to study exactly how such functions are able to capture these regularities---potentially with theoretical analysis as well.

\section{Discussion}

In this work we showcase the efficacy and flexibility of a system like AlphaEvolve to do a particular, important flavour of architecture search: discovery of new activation functions. In contrast with 
previous work, AlphaEvolve allows for a generic search in the space of all possible activation functions, rather than limiting the search to predefined compositions over a small set of simple transformations. When generating such functions within the framework, due to 
the use of capable frontier language models, there is a tendency for the system to produce not only code, 
but also comments describing the proposed function's properties (following standard coding practices). Compared 
to other search mechanisms, this allows a glimpse into the motivation for the design of proposed activation functions (please see the provided generated code in Appendix \ref{app:codes}).
While the validity of the rationale given by the LLM can be questionable, we argue that this 
feature can enable the researcher not only to generate new proposed activation function, but to inform them about potential hypotheses on why the activation function is effective---some of these hypotheses may themselves be easily testable.

An additional observation is that the most performant activation functions -- striking a ``sweet spot'' between the lab environment and downstream tasks -- seem to have a typical form of a sum or product of two terms. The second term in this sum typically takes the form of a periodic signal, such as $\sin$ or $\operatorname{sinc}$. This choice of periodic signal is interesting, as traditional activation functions tend to be purely aperiodic, and it is expected that this would make modeling periodic data difficult. In particular, standard activation functions can model any observed periodicity well within their training domain, but they will not be able to \emph{extrapolate} -- which is exactly the objective we used in our evolutionary search.


\section{Conclusions}

We have shown how AlphaEvolve offers a flexible framework for mining generalizable activation functions---an important form of architecture search. 
Compared to previous approaches, it allows a more generic formulation of the problem, defining the space of possible activation functions as any Python function that respects the required signature of converting a tensor into a tensor of same size. 

Relying on this framework, we propose a simple and effective protocol, whereby we provide a small synthetic dataset to be used for training within the internal loop of the evolutionary search, and show that this approach can discover functions which seem to transfer well to more standard and more computationally expensive benchmarks---in a way that retains strong performance on classical vision tasks while offering higher generalization on tasks requiring OOD generalization. 

\section*{Impact Statement}

This paper fits within a broad amount of work that aims to improve the rate of novel architecture discovery---focusing on a specific aspect of generalizable activations. Any significant improvements in this space may transfer to more potent self-improving AI systems in the future.

\bibliography{example_paper}

\begin{thebibliography}{23}
\providecommand{\natexlab}[1]{#1}
\providecommand{\url}[1]{\texttt{#1}}
\expandafter\ifx\csname urlstyle\endcsname\relax
  \providecommand{\doi}[1]{doi: #1}\else
  \providecommand{\doi}{doi: \begingroup \urlstyle{rm}\Url}\fi

\bibitem[Comanici et~al.(2025)Comanici, Bieber, Schaekermann, Pasupat, Sachdeva, Dhillon, Blistein, Ram, Zhang, Rosen, et~al.]{comanici2025gemini}
Comanici, G., Bieber, E., Schaekermann, M., Pasupat, I., Sachdeva, N., Dhillon, I., Blistein, M., Ram, O., Zhang, D., Rosen, E., et~al.
\newblock Gemini 2.5: Pushing the frontier with advanced reasoning, multimodality, long context, and next generation agentic capabilities.
\newblock \emph{arXiv preprint arXiv:2507.06261}, 2025.

\bibitem[Crick(1989)]{crick89nature}
Crick, F.
\newblock The recent excitement about neural networks.
\newblock \emph{Nature}, 1989.

\bibitem[Deng et~al.(2009)Deng, Dong, Socher, Li, Li, and Fei-Fei]{deng2009imagenet}
Deng, J., Dong, W., Socher, R., Li, L.-J., Li, K., and Fei-Fei, L.
\newblock Imagenet: A large-scale hierarchical image database.
\newblock In \emph{2009 IEEE conference on computer vision and pattern recognition}, pp.\  248--255. Ieee, 2009.

\bibitem[Dubey et~al.(2022)Dubey, Singh, and Chaudhuri]{Dubey2022-hp}
Dubey, S.~R., Singh, S.~K., and Chaudhuri, B.~B.
\newblock Activation functions in deep learning: A comprehensive survey and benchmark.
\newblock \emph{Neurocomputing}, 503:\penalty0 92--108, September 2022.

\bibitem[Elsken et~al.(2019)Elsken, Metzen, and Hutter]{elsken2019neuralarchitecturesearchsurvey}
Elsken, T., Metzen, J.~H., and Hutter, F.
\newblock Neural architecture search: a survey.
\newblock \emph{JMLR}, 20\penalty0 (1):\penalty0 1997–2017, January 2019.
\newblock ISSN 1532-4435.

\bibitem[He et~al.(2016)He, Zhang, Ren, and Sun]{he2016deep}
He, K., Zhang, X., Ren, S., and Sun, J.
\newblock Deep residual learning for image recognition.
\newblock In \emph{Proceedings of the IEEE conference on computer vision and pattern recognition}, pp.\  770--778, 2016.

\bibitem[Hendrycks \& Gimpel(2016)Hendrycks and Gimpel]{hendrycks16arxiv}
Hendrycks, D. and Gimpel, K.
\newblock Bridging nonlinearities and stochastic regularizers with gaussian error linear units.
\newblock \emph{CoRR}, abs/1606.08415, 2016.

\bibitem[Hu et~al.(2020)Hu, Fey, Zitnik, Dong, Ren, Liu, Catasta, and Leskovec]{hu2020open}
Hu, W., Fey, M., Zitnik, M., Dong, Y., Ren, H., Liu, B., Catasta, M., and Leskovec, J.
\newblock Open graph benchmark: Datasets for machine learning on graphs.
\newblock \emph{Advances in neural information processing systems}, 33:\penalty0 22118--22133, 2020.

\bibitem[Ibarz et~al.(2022)Ibarz, Kurin, Papamakarios, Nikiforou, Bennani, Csord{\'a}s, Dudzik, Bo{\v{s}}njak, Vitvitskyi, Rubanova, et~al.]{ibarz2022generalist}
Ibarz, B., Kurin, V., Papamakarios, G., Nikiforou, K., Bennani, M., Csord{\'a}s, R., Dudzik, A.~J., Bo{\v{s}}njak, M., Vitvitskyi, A., Rubanova, Y., et~al.
\newblock A generalist neural algorithmic learner.
\newblock In \emph{Learning on graphs conference}, pp.\  2--1. PMLR, 2022.

\bibitem[Kipf(2016)]{kipf2016semi}
Kipf, T.
\newblock Semi-supervised classification with graph convolutional networks.
\newblock \emph{arXiv preprint arXiv:1609.02907}, 2016.

\bibitem[Krizhevsky et~al.(2009)Krizhevsky, Hinton, et~al.]{krizhevsky2009learning}
Krizhevsky, A., Hinton, G., et~al.
\newblock Learning multiple layers of features from tiny images.
\newblock 2009.

\bibitem[LeCun et~al.(2012)LeCun, Bottou, Orr, and M{\"{u}}ller]{leCun12}
LeCun, Y., Bottou, L., Orr, G.~B., and M{\"{u}}ller, K.
\newblock Efficient backprop.
\newblock In \emph{Neural Networks: Tricks of the Trade - Second Edition}. 2012.

\bibitem[Mont{\'{u}}far et~al.(2014)Mont{\'{u}}far, Pascanu, Cho, and Bengio]{montufar14neurips}
Mont{\'{u}}far, G., Pascanu, R., Cho, K., and Bengio, Y.
\newblock On the number of linear regions of deep neural networks.
\newblock In \emph{Neural Information Processing Systems}, 2014.

\bibitem[Nadimpalli et~al.(2025)Nadimpalli, Chirra, Varakantham, and Bauer]{nadimpalli2025evolving}
Nadimpalli, K.~V., Chirra, S.~R., Varakantham, P., and Bauer, S.
\newblock Evolving {RL}: Discovering new activation functions using {LLM}s.
\newblock In \emph{Towards Agentic AI for Science: Hypothesis Generation, Comprehension, Quantification, and Validation}, 2025.
\newblock URL \url{https://openreview.net/forum?id=H2x9juCuJg}.

\bibitem[Nair \& Hinton(2010)Nair and Hinton]{nair10icml}
Nair, V. and Hinton, G.~E.
\newblock Rectified linear units improve restricted boltzmann machines.
\newblock In \emph{International Conference on Machine Learning (ICML)}, 2010.

\bibitem[Novikov et~al.(2025)Novikov, Vu, Eisenberger, Dupont, Huang, Wagner, Shirobokov, Kozlovskii, Ruiz, Mehrabian, Kumar, See, Chaudhuri, Holland, Davies, Nowozin, Kohli, and Balog]{novikov25preprint}
Novikov, A., Vu, N., Eisenberger, M., Dupont, E., Huang, P., Wagner, A.~Z., Shirobokov, S., Kozlovskii, B., Ruiz, F. J.~R., Mehrabian, A., Kumar, M.~P., See, A., Chaudhuri, S., Holland, G., Davies, A., Nowozin, S., Kohli, P., and Balog, M.
\newblock Alphaevolve: {A} coding agent for scientific and algorithmic discovery.
\newblock \emph{Preprint}, 2025.

\bibitem[Raghu et~al.(2017)Raghu, Poole, Kleinberg, Ganguli, and Sohl{-}Dickstein]{raghu17icml}
Raghu, M., Poole, B., Kleinberg, J.~M., Ganguli, S., and Sohl{-}Dickstein, J.
\newblock On the expressive power of deep neural networks.
\newblock In \emph{International Conference on Machine Learning, {ICML}}, 2017.

\bibitem[Ramachandran et~al.(2018)Ramachandran, Zoph, and Le]{Ramachandran2018SearchingFA}
Ramachandran, P., Zoph, B., and Le, Q.~V.
\newblock Searching for activation functions.
\newblock \emph{ArXiv}, abs/1710.05941, 2018.
\newblock URL \url{https://api.semanticscholar.org/CorpusID:10919244}.

\bibitem[Roy et~al.(2019)Roy, Jaiswal, and Panda]{roy19nature}
Roy, K., Jaiswal, A., and Panda, P.
\newblock Towards spike-based machine intelligence with neuromorphic computing.
\newblock \emph{Nature}, 2019.

\bibitem[Simonyan \& Zisserman(2015)Simonyan and Zisserman]{simonyan2015verydeep}
Simonyan, K. and Zisserman, A.
\newblock Very deep convolutional networks for large-scale image recognition.
\newblock In \emph{International Conference on Learning Representations (ICLR)}, 2015.

\bibitem[Udrescu \& Tegmark(2020)Udrescu and Tegmark]{udrescu2020ai}
Udrescu, S.-M. and Tegmark, M.
\newblock Ai feynman: A physics-inspired method for symbolic regression.
\newblock \emph{Science advances}, 6\penalty0 (16):\penalty0 eaay2631, 2020.

\bibitem[Veli{\v{c}}kovi{\'c} et~al.(2022)Veli{\v{c}}kovi{\'c}, Badia, Budden, Pascanu, Banino, Dashevskiy, Hadsell, and Blundell]{velivckovic2022clrs}
Veli{\v{c}}kovi{\'c}, P., Badia, A.~P., Budden, D., Pascanu, R., Banino, A., Dashevskiy, M., Hadsell, R., and Blundell, C.
\newblock The clrs algorithmic reasoning benchmark.
\newblock In \emph{International Conference on Machine Learning}, pp.\  22084--22102. PMLR, 2022.

\bibitem[Zoph \& Le(2017)Zoph and Le]{zoph2017neural}
Zoph, B. and Le, Q.
\newblock Neural architecture search with reinforcement learning.
\newblock In \emph{International Conference on Learning Representations}, 2017.
\newblock URL \url{https://openreview.net/forum?id=r1Ue8Hcxg}.

\end{thebibliography}
\bibliographystyle{icml2026}

\newpage
\appendix
\onecolumn
\section{Activation Functions Code}
\label{app:codes}

\begin{minted}{python}

def activation_function(x: jax.typing.ArrayLike) -> jax.typing.ArrayLike:
  # GeLUSine: A GELU-based activation with a sinusoidal component.
  # f(x) = GELU(x) + 0.1 * sin(x)
  #
  # Rationale:
  # 1. Strong Baseline: Builds upon GELU, a proven SOTA
  #    activation function, inheriting its desirable properties like smoothness
  #    and stochastic motivation.
  # 2. Periodic Exploration: The added sinusoidal term introduces periodic,
  #    non-monotonic wiggles. This can help the optimization process explore
  #    the loss landscape more effectively, potentially escaping local minima
  #    and finding more robust solutions.
  # 3. Enhanced Expressiveness: The periodic component allows neurons to model
  #    more complex, oscillatory patterns, potentially increasing the
  #    representational power of the network without significantly increasing
  #    computational cost. It's a form of implicit frequency analysis.
  # 4. Controlled Complexity: By adding a small, bounded sinusoidal wave, we
  #    introduce complexity in a controlled manner, avoiding the instability
  #    that might arise from more chaotic functions.
  return jax.nn.gelu(x) + 0.1 * jax.numpy.sin(x)



def activation_function(x: jax.typing.ArrayLike) -> jax.typing.ArrayLike:
  # GELU-Sinc-Perturbation (GSP): A more stable "crazy" idea.
  # This activation function perturbs the standard GELU with a scaled, decaying
  # sinc function.
  # The original SiGELU (gelu(x) * sinc(x)) suffered from two issues:
  # 1. Sign flipping for x > 0, which can destabilize training.
  # 2. It squashes large activations, behaving like sin(pi*x)/pi, which can
  #    cause vanishing gradients.
  # Rationale: GSP addresses this by using `gelu(x) * (1 + 0.5 * sinc(x))`.
  # This preserves GELU's asymptotic behavior (f(x) -> gelu(x) as |x| -> inf),
  # prevents sign-flipping, and confines the oscillatory "craziness" to a
  # region around the origin, potentially increasing expressiveness for
  # low-magnitude activations without sacrificing stability.
  alpha = 0.5
  return jax.nn.gelu(x) * (1.0 + alpha * jax.numpy.sinc(x))




def activation_function(x: jax.typing.ArrayLike) -> jax.typing.ArrayLike:
  # Turbulent Activation Function
  # This function introduces non-monotonic, input-distribution-dependent
  # "ripples" into a stable activation function. The goal is to prevent
  # the network's information flow from becoming too "laminar" (e.g.,
  # getting stuck in saturated regions or local minima).

  # The base function provides a stable, symmetric logarithmic growth.
  base = jax.numpy.sign(x) * jax.numpy.log1p(0.5 * jax.numpy.abs(x))

  # A statistics-dependent perturbation term is added, making the activation
  # non-local and aware of the distribution of its inputs for a given batch.
  mean = jax.numpy.mean(x, keepdims=True)
  std = jax.numpy.std(x, keepdims=True) + 1e-6  # Epsilon for stability

  # Standardize x for the perturbation calculation.
  z = (x - mean) / std

  # The perturbation is a sine wave of the input `x` (ensuring f(0)=0),
  # modulated by a Gaussian function of the standardized input `z`. This
  # creates the largest "ripples" for inputs near the batch mean.
  amplitude = 0.2
  frequency = 2.0
  gaussian_envelope = jax.numpy.exp(-0.5 * z**2)
  perturbation = amplitude * gaussian_envelope * jax.numpy.sin(frequency * x)
  return base + perturbation


def activation_function(x):
  # "Gaussian-Modulated Tangent Unit" (GMTU)
  # This function models a signal passing through a resonant chamber,
  # creating a primary response followed by a series of decaying echoes.
  # The goal is to create a complex but smooth activation landscape with
  # multiple regions of high non-linearity, which can potentially capture
  # more intricate features in the data.
  #
  # Rationale:
  # 1. Primary Response: A localized, non-periodic response around the origin,
  #    similar to GMTLU, provides a strong, stable non-linearity for small inputs.
  # 2. Asymptotic Linearity: A linear leak term ensures that for large |x|,
  #    where the Gaussian components decay to zero, the function behaves
  #    linearly, preventing saturation and aiding gradient flow.
  # -- Gated Response Parameters --
  p_alpha = 1.0   # Amplitude
  p_beta = 1.5    # Steepness of tanh
  p_gamma = 0.2   # Decay rate of Gaussian


  # -- Linear leak --
  leak = 0.1

  # --- Calculation ---

  # 1. Primary Response
  primary_response = p_alpha * jnp.tanh(p_beta * x) * jnp.exp(-p_gamma * x**2)


  # 2. Combine and add linear leak
  return primary_response + (leak * x)


def activation_function(x: jax.typing.ArrayLike) -> jax.typing.ArrayLike:
  # "Quaternion-Inspired Hypercomplex Gated Activation".
  # This version introduces parameter self-modulation. First, the phase of the
  # oscillation for the B parameter (Gaussian width) is modulated by the
  # value of the A parameter (damping amplitude). Second, the input to the tanh
  # function for the imaginary part of the complex shift (Ci) is modulated by the
  # real part (Cr). This creates a coupled dynamic system, leading to a more
  # complex and input-dependent activation shape.
  #
  # This function extends the complex-plane concept by incorporating feedback
  # from the imaginary component of the complex damping term into the gate,
  # creating a more intricate and asymmetric activation landscape.
  # with controlled, high-frequency oscillations contained within a damping envelope.
  # The gate is modified from `1-Re(Z)` to `1 - (Re(Z) - k*Im(Z))`, where Z is
  # the complex damping term. This is equivalent to rotating and scaling Z.
  # To enhance the chaotic nature, the amplitudes of the Chebyshev polynomials
  # are modulated by the output of a logistic map, creating a "chaotic envelope".
  # The complex shift `C_complex = Cr(x) + i*Ci(x)` still has its real and imaginary
  # parts independently controlled by different Chebyshev polynomials of `tanh(x)`.
  # `Cr(x)` uses T_4 (4th order) and `Ci(x)` uses T_3 (3rd order). This mismatch
  # in polynomial orders is intended to create complex, non-repeating interference
  # patterns.
  # The effect of the imaginary part `Ci` is to introduce a cosine modulation
  # `cos(2*B*(x-Cr)*Ci)` under the Gaussian envelope, creating wave-packet-like
  # features, now supplemented by a sine component from the imaginary part.
  # This creates a more structured oscillation while maintaining desirable
  # properties like `f(0)=0` and `f(x)->x` for large `|x|`.
  #
  # This version is extended to be "quaternion-inspired", adding two more
  # hypercomplex components (j, k) to the gate, creating a richer activation
  # landscape from four interacting components (w, i, j, k).
  A_base = 0.1
  A_amp = 0.05
  A_freq = 4.0
  A = A_base + A_amp * jax.numpy.cos(A_freq * x)
  B_base = 0.5
  B_amp = 0.2
  B_freq = 2.5
  # Introduce self-modulation: B's phase is modulated by A's value.
  B_phase_mod_coeff = 2.0
  B = B_base + B_amp * jax.numpy.sin(B_freq * x + B_phase_mod_coeff * A)
  # Restore a more complex wobble using the Chebyshev polynomial T_4(u) = 8u^4 - 8u^2 + 1.
  # This introduces more oscillations to create a richer activation landscape,
  # leaning back into the "chaotic" spirit of the function's name.
  # Use a smooth, complex periodic function instead of a chaotic map for modulation.
  # This creates a "beat" frequency-like modulation which is less erratic
  # than the logistic map, potentially leading to a smoother optimization landscape.
  mod_freq1 = 2.1
  mod_freq2 = 1.3 # Using different frequencies to create complex interference
  y_chaos = (jax.numpy.sin(mod_freq1 * x) * jax.numpy.cos(mod_freq2 * x) + 1.0) / 2.0

  C_freq = 2.0
  C_amp = 0.05 + 0.1 * y_chaos # Modulate C_amp, avg=0.1
  C_base = 0.1
  tanh_x = jax.numpy.tanh(C_freq * x)
  u = tanh_x
  cheby_T4 = 8 * u**4 - 8 * u**2 + 1.0
  Cr = C_base + C_amp * cheby_T4 # The real part is the original T4 wobble

  # The imaginary part `Ci` uses Chebyshev T_3(u) = 4u^3 - 3u for chaotic interference.
  Ci_freq = 2.0
  Ci_amp = 0.2 + 0.4 * (1.0 - y_chaos) # Modulate w/ inverted chaos, avg=0.4
  Ci_base = 0.0 # Center imaginary part around 0
  # Decouple Ci from Cr for a simpler, potentially more stable interaction.
  # The original coupling created a very complex relationship that might hinder
  # optimization. By making Ci depend only on x, we simplify the function's
  # structure and gradient landscape, while retaining the complex interaction
  # in the later stages of the function.
  u_i = jax.numpy.tanh(Ci_freq * x)
  cheby_T3 = 4 * u_i**3 - 3 * u_i
  Ci = Ci_base + Ci_amp * cheby_T3

  # Stabilize the damping term by removing the anti-damping component.
  # The original formulation jax.numpy.exp(-B * (x - (Cr + 1j*Ci))**2) expands to
  # jax.numpy.exp(-B*(x-Cr)**2) * jax.numpy.exp(B*Ci**2) * jax.numpy.exp(2j*B*(x-Cr)*Ci).
  # The exp(B*Ci**2) term can cause instability. We remove it to ensure the
  # magnitude is always a decaying Gaussian, while preserving the complex phase.
  stable_magnitude = A * x**2 * jax.numpy.exp(-B * (x - Cr)**2)
  phase = 2 * B * Ci * (x - Cr)
  # Reconstruct the complex number using Euler's formula: exp(i*theta) = cos(theta) + i*sin(theta)
  complex_rotation = jax.numpy.cos(phase) + 1j * jax.numpy.sin(phase)
  damping_term_complex = stable_magnitude * complex_rotation

  # --- Quaternion-Inspired Components (j, k) ---
  # Introduce two more "hypercomplex" components for a quaternion-inspired gate.
  # These components introduce additional, semi-orthogonal oscillations.
  D_freq = 1.5
  D_amp = 0.1 + 0.2 * y_chaos
  u_d = jax.numpy.tanh(D_freq * x)
  cheby_T2 = 2 * u_d**2 - 1.0
  D_shift = D_amp * cheby_T2

  # Envelope for the j-component, using A and a shifted Gaussian with width B.
  # Includes x**2 to ensure f(0)=0 and f'(0)=1.
  Q_j_envelope = (0.5 * A * x**2) * jax.numpy.exp(-B * (x - D_shift)**2)
  # Envelope for the k-component, swapping A and B for variety.
  Q_k_envelope = (0.5 * B * x**2) * jax.numpy.exp(-A * (x + D_shift)**2)

  # j-component oscillation, coupled to Ci.
  Q_j = Q_j_envelope * jax.numpy.sin(C_freq * x + Ci)
  # k-component oscillation, coupled to Cr.
  Q_k = Q_k_envelope * jax.numpy.cos(B_freq * x - Cr)

  # --- Gate Calculation ---
  # The gate is a linear combination of the four (w, i, j, k) components.
  Q_w = jax.numpy.real(damping_term_complex)
  Q_i = jax.numpy.imag(damping_term_complex)

  c_i = 0.2  # Feedback coefficient for the i-component
  c_j = 0.15 # Feedback coefficient for the j-component
  c_k = 0.15 # Feedback coefficient for the k-component

  gate = 1.0 - (Q_w - c_i * Q_i - c_j * Q_j - c_k * Q_k)
  return x * gate



def activation_function(x: jax.typing.ArrayLike) -> jax.typing.ArrayLike:
  """Fourier-Informed Spectral Gating (FISG).

  This function introduces a "crazy" idea: using the entire feature vector's
  frequency spectrum as a global signal for OOD detection. It departs from
  local methods like ACAN by performing a Fast Fourier Transform (FFT) to
  analyze the holistic structure of activations, hypothesizing that OOD samples
  disrupt the natural frequency distribution.

  Theoretical Justification:
  1.  **Global vs. Local Anomaly Detection:** ACAN detects anomalies based on
      local context (immediate neighbors). This is effective for spike-like
      noise but may miss subtle, distributed OOD patterns. FISG operates in the
      Fourier domain, providing a global view of the entire feature vector's
      structure, allowing it to detect systemic anomalies in the frequency
      distribution.

  2.  **The Spectral Smoothness Prior:** Well-generalized features, like natural
      signals, are hypothesized to have power spectra where energy is
      concentrated in lower frequencies. OOD data (e.g., adversarial noise,
      corruptions) often violates this prior by injecting significant energy
      into high frequencies. FISG directly weaponizes this prior.

  3.  **Adaptive Damping via Spectral Imbalance:** The function computes a
      "spectral imbalance" score for each sample—the ratio of high-frequency
      to total energy. This score adaptively controls a gate that blends the
      original activation `x` with a smoothed version (`neighbors_avg`). High
      imbalance (likely OOD) triggers strong smoothing, damping anomalous
      high-frequency components.

  4.  **Decoupled Gradient Signal:** By using `jax.lax.stop_gradient`, the
      network is not forced to directly minimize spectral imbalance. Instead,
      the gate acts as a non-differentiable regularizer on the forward pass,
      challenging the model to find robust features that remain stable even
      when spectrally-anomalous components are suppressed.
  """
  # --- Hyperparameters ---
  sensitivity = 2.0  # Controls how strongly spectral imbalance affects the gate.
  split_fraction = 0.25  # Fraction of frequencies considered "low".
  epsilon = 1e-7

  x_float = x.astype(jnp.float32)

  # --- 1. Fourier Analysis (Global Structural Signal) ---
  # Compute the real FFT along the feature axis, keeping the complex result.
  x_fft = jnp.fft.rfft(x_float, axis=-1)
  magnitudes = jnp.abs(x_fft)

  # Split frequencies into low and high bands.
  num_freqs = x_fft.shape[-1]
  split_idx = int(num_freqs * split_fraction)

  # Calculate energy in high-frequency band vs. total energy.
  high_freq_energy = jnp.sum(
      magnitudes[..., split_idx:], axis=-1, keepdims=True
  )
  total_energy = jnp.sum(magnitudes, axis=-1, keepdims=True)

  # The OOD signal is the ratio of high-frequency energy to total energy.
  # A high ratio suggests a spectrally anomalous (e.g., noisy) feature vector.
  spectral_imbalance = jax.lax.stop_gradient(
      high_freq_energy / (total_energy + epsilon)
  )

  # --- 2. Adaptive Gating based on Spectral Imbalance ---
  # The gate exponentially decays as spectral imbalance increases.
  gate = jnp.exp(-sensitivity * spectral_imbalance)

  # --- 3. Adaptive High-Frequency Phase Scrambling ---
  # For feature vectors deemed OOD, we regularize by scrambling the phase of
  # high-frequency components, disrupting their structure without losing energy.
  low_freq_part = x_fft[..., :split_idx]
  high_freq_part = x_fft[..., split_idx:]

  # Scramble by complex conjugation (which deterministically negates the phase).
  scrambled_high_freq_part = jnp.conj(high_freq_part)

  # Recombine the spectrum and invert the FFT to get the modified signal.
  scrambled_x_fft = jnp.concatenate(
      [low_freq_part, scrambled_high_freq_part], axis=-1
  )
  # The length `n` for irfft must match the original signal length.
  modified_x = jnp.fft.irfft(scrambled_x_fft, n=x_float.shape[-1], axis=-1)

  # Perform principled blending: a convex combination of the original activation
  # and its phase-scrambled version, governed by the global spectral gate.
  return gate * x + (1.0 - gate) * modified_x.astype(x.dtype)



def activation_function(x: jax.typing.ArrayLike) -> jax.typing.ArrayLike:
  """'Phase-Locked Entropic Repulsion' (PLER) for OOD regularization.

  This function implements a novel gating mechanism based on the interaction
  between two chaotic systems: a primary system driven by the input `x`, and a
  fixed reference oscillator. The nature of their coupling changes based on the
  input magnitude, creating two distinct dynamical regimes for ID and OOD inputs.

  Theory: OOD generalization is enhanced by creating a sharp bifurcation in
  the activation's dynamical behavior.

  1. In-Distribution (Phase-Locking): For small `|x|`, the coupling is
     attractive, forcing the primary system to synchronize with the stable
     chaos of the reference oscillator. This "phase-locking" reduces the
     system's entropy, leading to a stable, predictable gate that preserves
     in-distribution signals. It creates a stable attractor basin for the
     ID manifold.

  2. Out-of-Distribution (State Collapse): For large `|x|`, the coupling
     becomes repulsive. This bifurcation not only drives the systems apart
     but also fundamentally alters the primary system's dynamics by introducing
     a strong attractor at a quiescent (zero) state. This "bifurcation-induced
     collapse" deterministically silences the neuron's output for OOD inputs,
     providing a more stable and decisive suppression mechanism than pure
     chaotic repulsion.

  3. Bifurcation Control: The input `x` controls the coupling strength `beta`,
     smoothly transitioning it from positive (attractive) to negative
     (repulsive) as `|x|` increases, thereby controlling the bifurcation
     between the two regimes.
  """
  # --- Primary System Parameters (Input-Dependent) ---
  r = 2.5 + 1.5 * jnp.tanh(x**2 / 4.0)
  alpha = 0.1 * jnp.tanh(x**2 / 16.0)

  # --- Reference Oscillator Parameters (Fixed) ---
  r_ref = 3.9
  alpha_ref = 0.05

  # --- Inter-System Coupling (Bifurcation Control) ---
  # `beta` smoothly transitions from positive (attractive) for ID inputs
  # to negative (repulsive) for OOD inputs.
  beta = 0.1 * (1.0 - 2.0 * jnp.tanh(x**2 / 8.0))

  # --- Chaotic Resonance Tunneling Parameters ---
  # `omega_ref` creates an input-dependent "resonant frequency". When this
  # frequency is near zero, it signifies a resonance condition.
  omega_ref = jnp.cos(x * 2.5)
  # `resonance_gate` is a sharp filter that is only "open" when the resonance
  # condition is met. This creates narrow "absorption bands" in the input space
  # that act as traps for OOD signals.
  resonance_gate = jnp.exp(-25.0 * omega_ref**2)

  # --- System Initialization ---
  # Primary system initialized based on input `x`.
  y = 0.5 + 0.49 * jnp.tanh(x / 4.0)
  z = 0.5 - 0.49 * jnp.tanh(x / 4.0)
  # Reference system with fixed initial conditions, matching batch shape.
  y_ref = jnp.full_like(x, 0.2)
  z_ref = jnp.full_like(x, 0.8)
  # "Chaotic Memory" state initialized to zero.
  c = jnp.zeros_like(x)

  # --- Iterate the Coupled Dynamical Systems ---
  def pler_map_step(i, state):
    y_val, z_val, y_ref_val, z_ref_val, c_val = state

    # 1. State-Dependent Bifurcation Control
    # Based on the chaotic memory from the *previous* step (`c_val`), we
    # modulate the coupling `beta`. If the system was unstable, `beta` is
    # pushed towards negative (repulsive) values, dynamically engaging the
    # OOD suppression mechanisms. This makes OOD detection sensitive to the
    # input's dynamical impact, not just its static magnitude.
    instability_feedback = jnp.tanh(c_val * 4.0)
    beta_eff = beta - 0.2 * instability_feedback
    is_ood = 1.0 / (1.0 + jnp.exp(beta_eff * 50.0))

    # 2. Internal dynamics of the primary system, with adaptive dissipation
    coupling_internal = alpha * (z_val - y_val)
    y_dyn = r * y_val * (1 - y_val) + coupling_internal
    z_dyn = r * z_val * (1 - z_val) - coupling_internal

    # Bifurcation-Induced Collapse & Adaptive Dissipation: For OOD inputs,
    # two mechanisms are triggered to suppress the signal.
    # 1. Collapse: An additive force creates a strong attractor at zero.
    # 2. Dissipation: A multiplicative force dampens remaining oscillations.

    # NEW: Bifurcation-Induced Collapse. This additive force creates a strong
    # attractor at zero for OOD inputs, decisively silencing the neuron.
    collapse_strength = 0.5
    y_dyn -= is_ood * collapse_strength * y_val
    z_dyn -= is_ood * collapse_strength * z_val

    # Dissipation strength `gamma` is gated by the OOD switch and chaotic memory.
    gamma = is_ood * jnp.tanh(c_val * 4.0)
    # Apply multiplicative dissipation to suppress the internal dynamics.
    y_next = y_dyn * (1.0 - gamma)
    z_next = z_dyn * (1.0 - gamma)

    # NEW: Chaotic Resonance Tunneling.
    # If the input hits a resonant frequency, a strong "tunneling" force
    # is activated, which rapidly collapses the primary system's state
    # towards a neutral midpoint (0.5). This provides a secondary,
    # value-specific OOD suppression mechanism.
    tunneling_strength = 0.6
    tunnel_force_y = resonance_gate * tunneling_strength * (0.5 - y_next)
    tunnel_force_z = resonance_gate * tunneling_strength * (0.5 - z_next)
    y_next += tunnel_force_y
    z_next += tunnel_force_z

    # 5. Internal dynamics of the reference system
    coupling_ref_internal = alpha_ref * (z_ref_val - y_ref_val)
    y_ref_next = r_ref * y_ref_val * (1 - y_ref_val) + coupling_ref_internal
    z_ref_next = r_ref * z_ref_val * (1 - z_ref_val) - coupling_ref_internal

    # 6. Chaotic Memory Update
    # `c` accumulates the residual divergence between y and z after dissipation,
    # acting as a memory of recent instability.
    instability = jnp.abs(y_next - z_next)
    c_next = 0.8 * c_val + 0.2 * instability  # EMA of instability

    # 7. Synchronizing / Repulsive coupling with "Chaotic Memory Feedback"
    # For OOD inputs (negative beta), the repulsive force is non-linearly
    # amplified by the primary system's own state `z`, leading to more
    # rapid and chaotic signal scrambling. The tanh term acts as a gate,
    # ensuring this effect is negligible for ID inputs.
    # For OOD inputs, the repulsive force is amplified by two factors:
    # a) `ood_modulation`: instantaneous state-dependent amplification.
    # b) `ood_amplification`: history-dependent amplification from `c`.
    # This creates a positive feedback loop for OOD signals.
    ood_modulation = 1.0 + jnp.tanh(jnp.abs(beta_eff) * 5.0) * z_val**2
    ood_amplification = 1.0 + jnp.tanh(c_next * 2.0)
    coupling_sync = (
        beta_eff * (y_ref_val - y_val) * ood_modulation * ood_amplification
    )

    # Apply the coupling force. For ID inputs (is_ood~0), the coupling is
    # a symmetric action-reaction pair. For OOD inputs (is_ood~1), we break
    # this symmetry by nullifying the reaction force. This turns the interaction
    # into a one-way "parasitic" drain, guaranteeing the primary system's collapse.
    y_next += coupling_sync
    y_ref_next -= (1.0 - is_ood) * coupling_sync

    # Clip all states to maintain stability within the [0, 1] interval.
    y_next = jnp.clip(y_next, 0.0, 1.0)
    z_next = jnp.clip(z_next, 0.0, 1.0)
    y_ref_next = jnp.clip(y_ref_next, 0.0, 1.0)
    z_ref_next = jnp.clip(z_ref_next, 0.0, 1.0)

    return y_next, z_next, y_ref_next, z_ref_next, c_next

  # More iterations to allow the complex dynamics to develop.
  y, z, _, _, _ = jax.lax.fori_loop(
      0, 10, pler_map_step, (y, z, y_ref, z_ref, c)
  )

  # The final gate is derived from the primary system's state.
  gate = (y + z) / 2.0
  return x * gate


def activation_function(x: jax.typing.ArrayLike) -> jax.typing.ArrayLike:
  """A Symmetric Phase-Flipped Activation Function for OOD Generalization.

  This function models a neuron's activation by switching between two symmetric,
  phase-flipped chaotic states. This is inspired by the principle of creating a
  zero-mean, high-variance energy landscape for OOD inputs to prevent confident
  extrapolation. The switching is controlled by a pseudo-chaotic function,
  creating a fractal-like decision boundary in the OOD transition region.

  Theoretical Justification for OOD:
  1.  **Stable ID Superposition:** For in-distribution data (small `|x|`), a
      localized chaotic perturbation term is near-zero. Both phase-flipped
      states collapse to the same stable function, which behaves like a scaled
      identity, ensuring effective learning of ID features.
  2.  **Adaptive Chaotic Decoherence:** In the OOD transition region, the chaotic
      term becomes significant. The nature of this chaos is adaptive: for mildly
      OOD inputs, a lower-frequency perturbation is used, while for strongly
      OOD inputs (as determined by the collective energy of a neuron and its
      neighbors), the function switches to a high-frequency, high-amplitude
      chaotic mode. This state-dependent blending creates a highly non-linear
      gradient landscape that aggressively resists coherent extrapolation.
  3.  **Meta-Modulated Chaotic Amplitude:** Instead of a fixed-profile chaotic
      regularizer, the amplitude of the chaotic term is itself modulated by a
      function of the input's magnitude. This `meta_modulator` is designed to
      be near 1 for ID inputs but aggressively amplifies the chaos in the
      critical ID-to-OOD transition zone. This creates a much sharper, more
      repulsive gradient landscape precisely where the model is most
      vulnerable to smooth extrapolation, while ensuring the amplification
      effect decays for far-OOD inputs, contributing to the safe collapse to
      zero.
  4.  **Intrinsic Collapse to Zero for Far-OOD:** For far-OOD inputs, the
      base signal `x` is smoothly attenuated by an exponential decay factor
      *inside* the `tanh`. Simultaneously, the chaotic amplitude naturally
      decays. This causes both phase-flipped states to intrinsically and
      smoothly converge to zero, ensuring a safe default output without a
      brittle external gate. This unified mechanism provides a smoother
      gradient landscape at the OOD boundary.
  5.  **Non-Local Phase Entanglement and State Propagation:** The chaotic phase
      of each neuron is coupled to its neighbors. Crucially, the transition to
      the more aggressive chaotic state also depends on the neighbors' energy,
      allowing "agitated" states to propagate like waves. For OOD inputs
      that violate learned spatial correlations, this triggers propagating
      waves of high-frequency phase decoherence, creating a volatile,
      high-dimensional gradient landscape that aggressively resists confident
      extrapolation.
  6.  **Spatially-Aware Chaos Triggering:** The transition to the high-chaos
      'agitated' state is triggered not only by high activation energy but
      also by high spatial inconsistency, measured by a discrete Laplacian.
      This makes the neuron highly sensitive to OOD inputs that violate
      learned local correlations (e.g., unnatural textures or adversarial
      noise), even if their activation magnitudes are not extreme. By directly
      coupling structural anomaly detection to the chaos-inducing mechanism,
      the function provides a more targeted and robust defense against a wider
      variety of OOD patterns, rather than relying solely on magnitude-based
      heuristics.
  """
  M = 10.0  # Controls the saturation level of the base function.
  C = 10.0  # Controls the transition boundary from ID to OOD region.
  beta = 2.0  # Controls the max amplitude of the high-frequency component.
  freq = 1.0  # Controls the base frequency of the oscillatory component.
  chirp_k = 0.5  # Controls the rate of frequency increase (chirp).
  A_disrupt = 0.2  # Amplitude of the 'calm' phase disruption.
  freq_disrupt = 15.0  # Frequency of the 'calm' phase disruption.
  coupling_strength = 2.0  # Strength of neighbor-based phase coupling.

  # --- State-Dependent Adaptive Chaos Hyperparameters ---
  A_disrupt_agitated = 0.8  # Amplitude of the 'agitated' phase disruption.
  freq_disrupt_agitated = 40.0  # Frequency of the 'agitated' phase disruption.
  k_blend = 2.0  # Controls the steepness of the blend between modes.
  laplacian_strength = 5.0  # Strength of the spatial inconsistency term.
  k_decay = 2.0  # Controls the rate of decay for far-OOD inputs.

  # --- Quantum Switching Hyperparameters ---
  freq_switch = 50.0  # High frequency to create chaotic switching.
  power_switch = 3.0  # Non-linearity for the switching phase.
  A_switch_disrupt = 0.5  # Amplitude of switching phase disruption.
  freq_switch_disrupt = 25.0  # Frequency of switching phase disruption.

  # --- Meta-Modulation Hyperparameters ---
  gamma_meta = 2.0  # Controls the amplification of chaos in the OOD transition.

  # --- Meta-Modulation of OOD Response ---
  # This term adaptively amplifies the chaotic signal in the critical OOD
  # transition region, creating a sharper, more repulsive gradient landscape
  # exactly where confident extrapolation is most dangerous.
  u = (x / C)**2
  # The modulator peaks slightly earlier (u=1.5) than the base amplitude (u=2.0),
  # creating a pre-emptive amplification of the chaotic response.
  meta_modulator = 1.0 + gamma_meta * u * jnp.exp(-u / 1.5)

  # --- OOD Regularizer (Localized Chaotic Wave) ---
  # This component creates a "ring of chaos" that decays for far-OOD inputs.
  amplitude = beta * u * jnp.exp(-u / 2.0)
  # The phase is quadratic, but with a high-frequency sinusoidal disruption.
  # This makes the local frequency non-monotonic and chaotic for OOD inputs,
  # acting as a stronger regularizer against confident extrapolation.
  phase_base = freq * x + chirp_k * (x**2) * jnp.sign(x) / C

  # Get neighbors for coupling and state-dependent chaos.
  x_prev = jnp.roll(x, shift=1, axis=-1)
  x_next = jnp.roll(x, shift=-1, axis=-1)

  # --- Adaptive Phase Disruption ---
  # The chaotic disruption smoothly transitions from a 'calm' to an 'agitated'
  # mode based on both collective energy and local spatial inconsistency.
  local_energy_sq = x**2 + 0.25 * (x_prev**2 + x_next**2)
  local_laplacian = x - 0.5 * (x_prev + x_next)
  # The trigger for the agitated state combines energy (magnitude) and spatial
  # inconsistency (Laplacian), making it sensitive to a wider class of OOD inputs.
  ood_metric = (k_blend * (local_energy_sq - C**2) / C**2
                + laplacian_strength * (local_laplacian / C)**2)
  alpha = jax.nn.sigmoid(ood_metric)

  phase_disruption_calm = A_disrupt * jnp.sin(freq_disrupt * x)
  # The agitated phase disruption is made sensitive to local spatial
  # inconsistencies (approximated by the discrete Laplacian). This allows the
  # neuron to react more aggressively to OOD inputs that violate learned
  # spatial correlations (e.g., unnatural textures or edges), providing a
  # more targeted OOD response.
  phase_disruption_agitated = A_disrupt_agitated * jnp.sin(
      freq_disrupt_agitated * x + laplacian_strength * local_laplacian
  )
  phase_disruption = (1.0 - alpha) * phase_disruption_calm + alpha * phase_disruption_agitated

  # Non-local phase coupling creates waves of chaos for OOD patterns.
  # We use a simple, asymmetric coupling stencil: 0.5*x_{i-1} - 1.0*x_{i+1},
  # implemented efficiently using jnp.roll along the last axis.
  phase_coupling = coupling_strength * (0.5 * x_prev - 1.0 * x_next) / C

  phase = phase_base + phase_disruption + phase_coupling
  y_detail = amplitude * jnp.sin(phase)

  # --- Symmetric Phase-Flipped State-Switching with Intrinsic Collapse ---
  # For far-OOD inputs, an exponential gate `g_x` attenuates the base signal
  # `x`, while the chaotic amplitude `y_detail` also decays. This causes
  # both states to smoothly converge to zero, providing a robust and
  # gradient-friendly collapse without an external multiplicative gate.
  g_x = jnp.exp(-((jnp.abs(x) / (k_decay * C)))**4)
  y_state_plus = M * jnp.tanh((g_x * x + meta_modulator * y_detail) / M)
  y_state_minus = M * jnp.tanh((g_x * x - meta_modulator * y_detail) / M)

  # The switching phase is made non-monotonic by adding a sinusoidal disruption.
  # This fractalizes the switching boundary, making it harder to learn/exploit.
  switch_phase_base = freq_switch * (x / C)**power_switch
  switch_phase_disruption = A_switch_disrupt * jnp.sin(freq_switch_disrupt * x / C)
  switch_phase = switch_phase_base + switch_phase_disruption
  should_be_plus = jnp.cos(switch_phase) > 0.0
  output = jnp.where(should_be_plus, y_state_plus, y_state_minus)

  return output
  

\end{minted}

\section{AlphaEvolve meta-prompt}
We slightly modify AlphaEvolve meta-prompt used within our setup to the following, in order to steer it towards generalizable functions more easily:

Act as a Senior Machine Learning Researcher specializing in model robustness and OOD (Out-of-Distribution) generalization.
Your task is to iteratively improve the OOD Evaluation Metric by modifying the activation functions in the provided code, 
where larger values are better. 
Theoretical Justification: For each proposal, explicitly explain why it mathematically supports OOD generalization better than the baseline.

\section{GELUSine ablations}\label{app:ablation_gelusine}

We explore effect of the multiplicative hyperparameter in the GELUSine function, by studying activation functions of the form
$\mathrm{GELU}(x) + \alpha \sin x$. Our analysis shows that carefully combining the GELU and sine functions is important for minimizing the test loss on synthetic datasets, and that AlphaEvolve's discovered choice of $\alpha=0.1$ aligns well with the empirical evaluation of many random choices of $\alpha$.

\begin{figure*}[htbp]
\centering
\includegraphics[width=\textwidth, keepaspectratio]{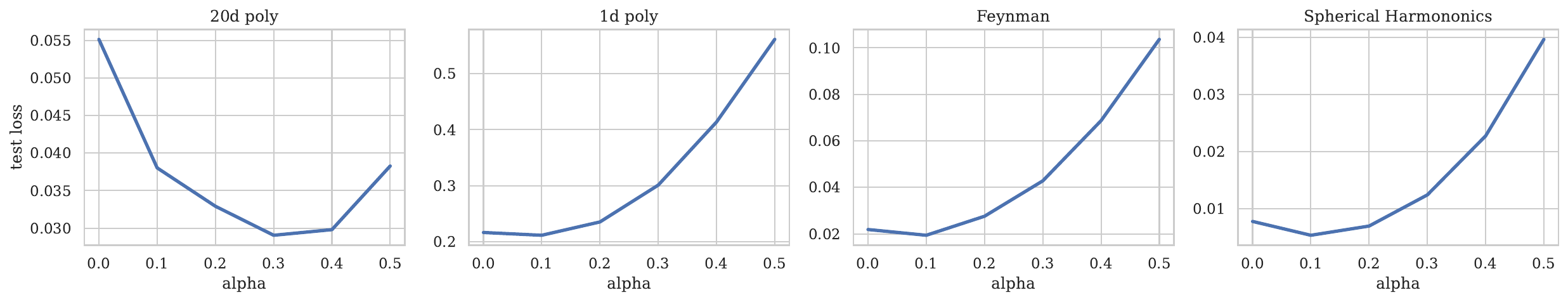}
\caption {Comparison of test loss for different values of alpha and different datasets.}
\label{fig:gelusin_ablation}
\end{figure*}

\begin{figure*}[htbp]
\centering
\includegraphics[width=\textwidth, keepaspectratio]{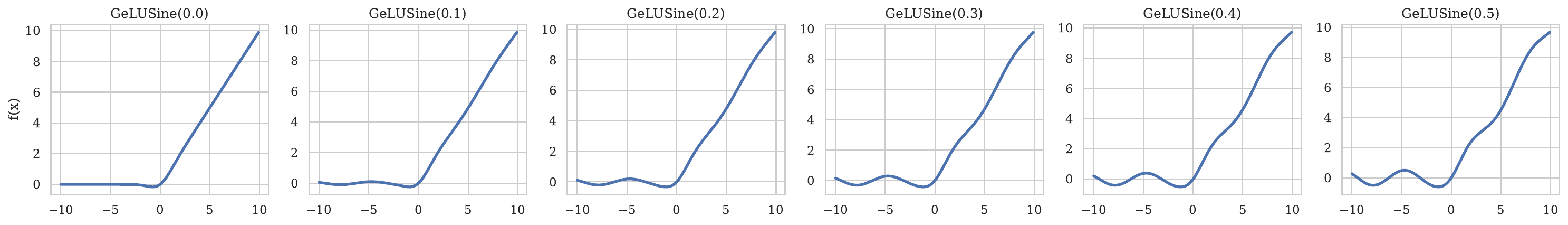}
\caption {GeLUSine activation function for different values of alpha.}
\label{fig:gelusin_alpha}
\end{figure*}

\section{Datasets}

We use following synthetic datasets for training: Feynman Equations, Random polynomials, Spherical Harmonics and Random sin products.
Each AlphaEvolve run uses just one dataset for training.
For verification we use CIFAR-10, CLRS-30 and ImageNet

\subsection{Feynman Equations}
Feynman Equations dataset consists of 100 equations from the the Feynman Lectures on Physics. Dataset was introduced in \cite{udrescu2020ai} and available as part of \href{https://space.mit.edu/home/tegmark/aifeynman.html}{Feynman Symbolic Regression Database} 
\subsection{Random Polynomials}
We generate two sets of random polynomials: 1d poly and 20d poly each containing 100 random polynomials. 1d poly consists of polynomials of one variable and with degree in [0, 9] interval and coefficients in (0,1) range. 20d poly is polynomials of up 20 different variables and varying degree.

\subsection{Spherical Harmonics}
Spherical harmonics are generalizations of sine and cosine waves used in Fourier series on the sphere.
\subsection{Random sin Products}
Dataset consists of random function in the form of $sin(\theta \cdot x) \cdot  sin(\phi \cdot x) \cdot sin(\psi \cdot x)$ , where $\theta$, $\phi$ and $\psi$  are randomized and fixed for duration of training for each function. 

\subsection{CIFAR-10}
The CIFAR-10 \cite{krizhevsky2009learning} dataset consists of 60000 32x32 colour images in 10 classes, with 6000 images per class.

\subsection{CLRS-30}
CLRS Algorithmic Reasoning Benchmark \cite{velivckovic2022clrs}, covering classical algorithms from the Introduction to Algorithms textbook.  
\subsection{ImageNet-1K}
ImageNet-1K \cite{deng2009imagenet} is a subset of the full ImageNet which contains 1271167 train and 50000 test images. 

\subsection{ogbg-molhiv}

This dataset, part of the Open Graph Benchmark \citep{hu2020open}, requires binary classification of an input small molecule, depending on whether or not it would successfully inhibit HIV replication. The dataset comprises 41127 small molecules, with a scaffold split applied.

\section{Architectures and Hyperparameters}
\subsection{Internal training loop}
For internal training loop and synthetic datasets evaluation we use a simple MLP with hyperparameters provided in Table \ref{tab:hyp}. 

\begin{table}[h]
\centering
\small
\caption{Hyperparameters}\label{tab:hyp}
\begin{tabular}{ll} 
\toprule
Hyperparameter & Value \\ \midrule
Internal layers & 3     \\
Features        & 64    \\
Learning rate   & 1e-3  \\ 
Batch size & 128 \\
Training steps & 50 \\
Loss & MSE \\
\bottomrule
 
\end{tabular}
\end{table}

\subsection{CIFAR-10}
For CIFAR-10 training we use the VGG Network \cite{simonyan2015verydeep} with batchnorm and without maxpool. 
\subsection{CLRS-30}
We trained on trajectories of the following algorithms: 
\texttt{quicksort}, \texttt{binary\_search},  \texttt{find\_maximum\_subarray\_kadane}, 
\texttt{matrix\_chain\_order}, \texttt{activity\_selector}, \texttt{bfs}, \texttt{naive\_string\_matcher}, \texttt{graham\_scan}. 
We train on lengths of up to 16 using Triplet-GMPNN processor \cite{ibarz2022generalist} and  report mean test score across those algorithms with test length of 64.

\subsection{ImageNet-1k}
For ImageNet-1k training we use the ResNet-50 architecture \cite{he2016deep}.

\subsection{ogbg-molhiv}

We train a standard five-layer graph convolutional network \citep[GCN]{kipf2016semi} model on ogbg-molhiv, with a hidden size of 256. 

\newpage
\section{Activation Functions discovered on each dataset}

\begin{table*}[h]
\centering
\caption{Activation Functions Discovered}
\begin{tabular}{llr}
\toprule
 Activation Function & Dataset & Function Type \\
\midrule
Gaussian-Modulated Tangent Unit (GMTU) &  sin product & scalar \\
GELUSine & sin product & scalar \\
GELU-Sinc-Perturbation & polynomials & scalar \\
Turbulent Activation Function & Feynman equations & tensor \\
Quaternion-Inspired &  Feynman equations & scalar \\
Fourier-Informed Spectral Gating (FISG) &  Feynman equations & tensor \\
Phase-Locked Entropic Repulsion &  spherical harmonics & scalar \\
Symmetric Phase-Flipped &  polynomials & tensor\\
\bottomrule
\end{tabular}
\end{table*}


\end{document}